\documentclass[sigconf,nonacm]{acmart}

\usepackage[ruled,vlined]{algorithm2e}
\usepackage{multirow}
\usepackage{enumitem}
\PassOptionsToPackage{hyphens}{url}\usepackage{hyperref}
\usepackage{xurl}

\AtBeginDocument{
  \providecommand\BibTeX{{
    \normalfont B\kern-0.5em{\scshape i\kern-0.25em b}\kern-0.8em\TeX}}}

\makeatletter
\newcommand{\removelatexerror}{\let\@latex@error\@gobble}
\makeatother

\begin{document}

\title{LightAutoML: AutoML Solution for a Large Financial Services Ecosystem}

\author{Anton Vakhrushev}
\affiliation{
  \institution{Sber AI Lab}
   }
\email{AGVakhrushev@sberbank.ru}

\author{Alexander Ryzhkov}
\affiliation{
  \institution{Sber AI Lab}
  }
\email{AMRyzhkov@sberbank.ru}

\author{Maxim Savchenko}
\affiliation{
  \institution{Sber AI Lab}
  }
\email{MSSavchenko@sberbank.ru}

\author{Dmitry Simakov}
\affiliation{
  \institution{Sber AI Lab}
  }
\email{Simakov.D.E@sberbank.ru}

\author{Rinchin Damdinov}
\affiliation{
  \institution{Sber AI Lab}
  }
\email{RGDamdinov@sberbank.ru}

\author{Alexander Tuzhilin}
\affiliation{
  \institution{Stern School of Business, NYU}
  }
\email{atuzhili@stern.nyu.com}

\newcommand{\GG}[1]{\textcolor{red}{[GG: #1]}}
\renewcommand{\shortauthors}{Vakhrushev and Ryzhkov, et al.}

\begin{abstract}
We present an AutoML system called LightAutoML developed for a large European financial services company and its ecosystem satisfying the set of idiosyncratic requirements that this ecosystem  has for AutoML solutions. Our framework was piloted and deployed in numerous  applications and performed at the level of experienced data scientists while building high-quality ML models significantly faster than these data scientists. We also compare the performance of our system with various general-purpose open source AutoML solutions and show that it performs better for most of the ecosystem and OpenML problems. We also present the lessons we learned while developing the AutoML system and moving it into production.
\end{abstract}

\keywords{AutoML, financial modeling, banking applications, deployment}

\settopmatter{printfolios=true}
\maketitle

\section{Introduction}
\label{sec:intro}

AutoML has attracted much attention over the last few years, both in the industry and academia~\cite{guyon2019analysis}. 
In particular, several companies, such as H2O.ai\footnote{\url{https://www.h2o.ai/}}, DataRobot~\footnote{\url{https://www.datarobot.com/}}, DarwinAI~\footnote{\url{https://darwinai.ca/}}, and OneClick.ai\footnote{\url{https://www.oneclick.ai/}}, and existing AutoML libraries, such as AutoWeka~\cite{thornton2013auto, kotthoff2017auto}, MLBox, AutoKeras~\cite{jin2019auto}, Google's Cloud AutoML\footnote{\url{https://cloud.google.com/automl/}}, Amazon's AutoGluon~\cite{erickson2020autogluon}, IBM Watson AutoAI \footnote{\url{https://www.ibm.com/cloud/watson-studio/autoai}}, and Microsoft Azure AutoML\footnote{\url{https://docs.microsoft.com/en-us/azure/machine-learning/concept-automated-ml}} have provided industrial solutions that automatically build ML-based models. Most of these libraries propose general-purpose AutoML solutions that automatically develop ML-based models across a broad class of applications in financial services, healthcare, advertising, manufacturing, and other industries \cite{guyon2019analysis}. 
\par
The key assumption of this horizontal approach is that the process of automated model development remains the same across all these applications.
In this paper, however, we focus on developing a \emph{vertical} AutoML solution suitable for the needs of the ecosystem \cite{pidun2019you} of a large European financial services company comprising a wide range of banking and other types of financial as well as non-financial services, including telecommunications, transportation, and e-commerce for the B2B and B2C sectors of the economy.
We argue in the paper that such ecosystem has an idiosyncratic set of requirements for building ML models and would be better served by a domain-specific AutoML solution rather than using a generic horizontal AutoML system.
In particular, our ecosystem has the following set of requirements:
\begin{itemize}[leftmargin=*]
    \item AutoML system should be able to work with different types of data collected from hundreds of different information systems. Moreover, often such AutoML system changes more rapidly than it can be fully documented and painstakingly preprocessed by data scientists for the ML tasks using ETL tools.
    \item Many of our models are typically built on large datasets, having thousands or tens of thousands of features and millions of records. This makes it important to develop fast AutoML methods that efficiently handle such datasets.
     \item The number of production-level models across our complex ecosystem is very large, measured in thousands, and continues to increase rapidly, forming a Long Tail in terms of their popularity and economic efficiency. This makes it necessary for the AutoML system to accurately build and maintain all these models efficiently and cost-effectively. Furthermore, besides building production models, it is necessary to build a large number of models to validate numerous hypotheses tested across the entire ecosystem and do it efficiently.
    \item Many of our business processes are non-stationary and are rapidly changing over time which complicates the process of validating and keeping up to date the ML models that are included in these evolving processes. This results in the need to satisfy specific model validation needs, including out-of-time validation and validation of client behavioral models (models that take a sequence of single object states as input).
\end{itemize}
\par
In this paper, we introduce a vertical type of AutoML, called \emph{LightAutoML}, which focuses on the aforementioned needs of our complex ecosystem and that has the following characteristics. 
First, it provides nearly optimal and fast hyperparameter search, but it does not optimize them directly, nevertheless ensuring satisfactory~\cite{simon1956rational} results.
Furthermore, we dynamically keep the balance between hyperparameter optimization and speed, making sure that our solutions are optimal on small problems and fast enough on large ones.
Second, we purposely limit the range of ML models to only two types, i.e., gradient boosted decision trees (GBMs) and linear models, instead of having large ensembles of multiple algorithms.  It is done to speed up LightAutoML execution time without sacrificing its performance for our types of problems and data. 
Third, we present a unique method of choosing preprocessing schemes for different features used in our models based on certain types of meta-statistics and selection rules.
\par
We tested the proposed LightAutoML system on a wide range of open and proprietary data sources across a wide range of applications and demonstrated its superior performance in the experiments. 
Furthermore, we deployed LightAutoML in our ecosystem in numerous applications across five different platforms which enabled the company to save millions of dollars and present our experiences with this deployment and business outcomes. 
In particular, the initial economic effects of LightAutoML in these applications range from 3\% to 5\% of the total ML economic effects from deployed ML solutions in the company.
Moreover, LightAutoML provided certain novel capabilities that humans cannot perform, such as generating massive amounts of ML models in record time in the non-stop (24-7-365) working mode. 
\par
In this paper, we make the following contributions. First, we present the LightAutoML system developed for the ecosystem of a large financial services company comprising a wide range of banking and other types of financial and non-financial services.
Second, we compare LightAutoML with the leading general-purpose AutoML solutions and demonstrate that LightAutoML outperforms them across several ecosystem applications and on the open source AutoML benchmark \emph{OpenML} \cite{gijsbers2019open}. 
Third, we compare the performance of LightAutoML models with those manually tuned by the data scientists and demonstrate that LightAutoML models usually outperform data scientists. 
Finally, we describe our experiences with the deployment of LightAutoML in our ecosystem. 

\section{Related work}

The early work on AutoML goes back to the mid-’90s when the first papers on hyperparameter optimization were published~ \cite{king1995statlog, kohavi1995automatic}. Subsequently, the concepts of AutoML were expanded and the interest in AutoML significantly increased after the publication of the Auto-WEKA paper \cite{thornton2013auto} in 2013 and the organization of the AutoML workshop at ICML in 2014.
\par
One of the main areas of AutoML is the problem of hyperparameter search, where the best performing hyperparameters for a particular ML model are determined in a large hyperparameter space using various optimization methods  \cite{bergstra2011implementations}. Another approach is to estimate the probability that a particular hyperparameter is the optimal one for a given model using Bayesian methods that typically use historical data from other datasets and the previously estimated models on these datasets  \cite{swersky2013multi, springenberg2016bayesian, perrone2017multiple}.  Other methods, besides the hyperparameter optimization, try to select the best models from the space of several possible modeling alternatives  
\cite{yang2019oboe, lee2018deep, olson2019tpot, santos2019visus}. For example, TPOT \cite{olson2019tpot} generates a set of best-performing models from Sklearn and XGboost and automatically chooses the best subset of models. Further, other papers focus on the problem of automated deep learning model selection and optimization  \cite{lee2018deep, zimmer2020auto}. Finally, several papers propose various methods of automatic feature generation \cite{olson2019tpot}.
\par
In addition to the specific approaches to AutoML described above, there has been a discussion in the AutoML community on what AutoML is and how to properly define it, with different authors expressing their points of view on the subject. In particular, while some approaches focus only on the modeling stage of the CRISP-DM model lifecycle, other approaches take a broader view of the process and cover other lifecycle stages. For example, according to Shubha Nabar from Salesforce, \textquotedblleft most auto-ML solutions today are either focused very narrowly on a small piece of the entire machine learning workflow, or are built for unstructured, homogenous data for images, voice and language\textquotedblright\  \cite{Nabar}. Then she argues that the real goal of the AutoML system is the end-to-end approach across the whole CRISP-DM cycle that \textquotedblleft transforms customer data into meaningful actionable predictions\textquotedblright\ \cite{Nabar}. A similarly broad view of AutoML is presented in  \cite{guyon2003introduction} where it was argued that AutoML focuses on \textquotedblleft removing the need for human interaction in applying machine learning (ML) to practical problems\textquotedblright.
A similar argument for a broader view of AutoML as an end-to-end process taking the input data and automatically producing an optimized predictive model was presented in \cite{Janakiram}. 
\par
Furthermore, some successful examples of industry-specific AutoML solutions for medical, financial, and advertisement domains are reviewed in  \cite{guyon2019analysis}. One particular application of AutoML in the financial sector is presented in  \cite{agrapetidou2020automl}, where a simple general-purpose AutoML system, utilizing models of random forest, support vector machines, k-nearest neighbors with different kernels, and other ML methods, is created for the task of detecting bank failures. This system is an experimental proof-of-concept prototype focusing on a narrow task of bank failures but not an industrial-level AutoML solution designed for a broad class of financial applications.
\par
In this paper, we focus on a broader approach to AutoML, which includes the stages of data processing, model selection, and hyperparameter tuning. This is in line with other popular approaches to AutoML incorporated into systems such as AutoGluon \cite{erickson2020autogluon}, H2O \cite{ledell2020h2o}, AutoWeka \cite{thornton2013auto}, TPOT \cite{olson2019tpot}, Auto-keras \cite{jin2019auto}, AutoXGBoost \cite{thomas2018automatic}, Auto-sklearn \cite{feurer2019auto}, Amazon SageMaker \cite{das2020amazon}.

\section{Overview of LightAutoML}
\label{sec:lama}

In this section, we describe 
implementation details of LightAutoML, an open source modular AutoML framework that can be accessed at our GitHub repository\footnote{\url{https://github.com/sberbank-ai-lab/LightAutoML}}.
LightAutoML consists of modules that we call \emph{Presets}. They are focused on the end-to-end model development for typical ML tasks. 
Currently, LightAutoML supports the following four Preset modules.  First, \emph{TabularAutoML} Preset 
focuses on classical ML problems defined on tabular datasets. 
Second, \emph{WhiteBox} Preset solves binary classification tasks on tabular data using simple interpretable algorithms, such as Logistic Regression over discretized features and \emph{Weight of Evidence (WoE)} \cite{zeng2014necessary} encoding. This is a commonly used approach to model the probability of a client default in banking applications because of interpretability constraints posed by regulators and high costs of loan approval for bad customers.
Third, \emph{NLP} Preset is the same as Tabular but is also able to combine tabular pipelines with the NLP tools, such as specific feature extractors or pre-trained deep learning models. 
The last \emph{CV} Preset implements some basic tools to work with image data.
In addition, it is also possible to build custom modules and Presets using LightAutoML API. Some examples are also available at our GitHub repository and on Kaggle Kernels\footnote{\url{https://www.kaggle.com/simakov/lama-custom-automl-pipeline-example}}.
Although LightAutoML supports all four Presets, only TabularAutoML is currently used in our production-level system. Therefore, 
we will focus on it in the rest of this paper.
\par
\begin{figure}[htbp]
  \centering
  \includegraphics[width=\linewidth]{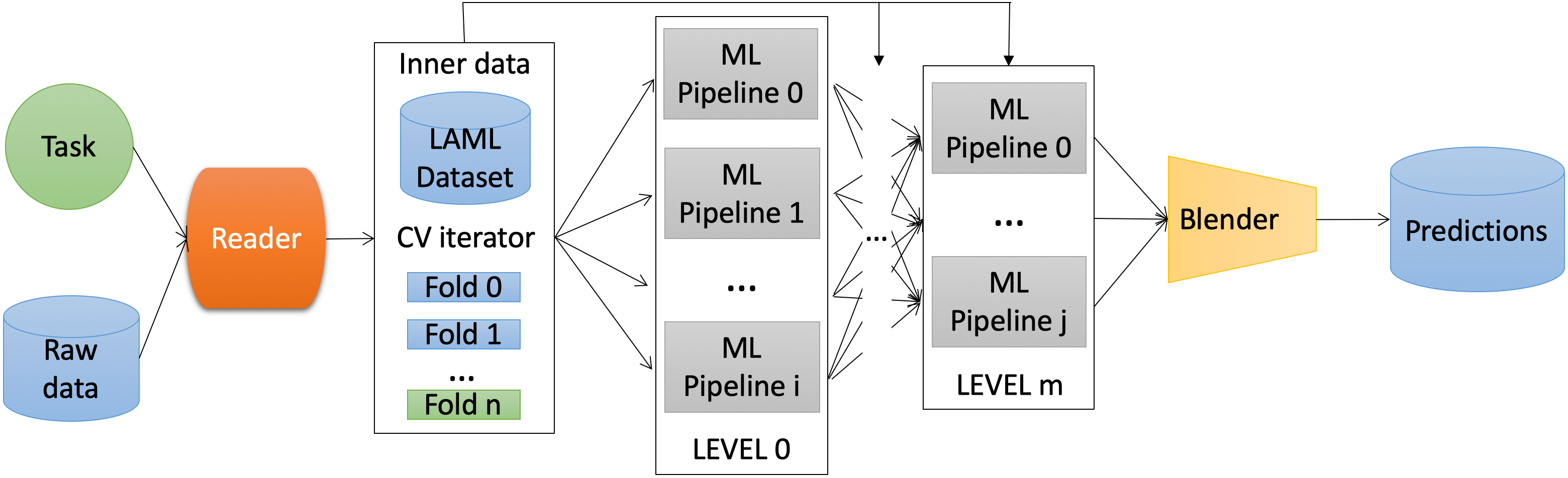}
  \caption{Main components of LightAutoML Pipeline}
  \Description{Smth2}
  \label{LamaBig}
\end{figure}

A typical LightAutoML pipeline scheme is presented
in Figure~\ref{LamaBig}. Each pipeline contains:
\begin{itemize}[leftmargin=*]
    \item \emph{Reader}: an object that receives raw data and task type as input, calculates some useful metadata, performs initial data cleaning, and decides what data manipulations should be done before fitting different models.
    \item \emph{LightAutoML inner datasets} which contain metadata and CV iterators that implement validation schemes for the datasets.
    \item \emph{Multiple ML Pipelines} that are stacked \cite{ting1997stacking} and/or blended (averaged) via \emph{Blender} to get a single prediction.
\end{itemize}

\par
An ML pipeline in LightAutoML is one or multiple ML models that share a single data preprocessing and validation scheme. The preprocessing step may have up to two feature selection steps, a feature engineering step or may be empty if no preprocessing is needed. 
The ML pipelines can be computed independently on the same datasets and then blended together using averaging (or weighted averaging). Alternatively, a stacking ensemble scheme can be used to build multilevel ensemble architectures.
\par
In the next section we describe TabularAutoML Preset and compare it with other popular open source AutoML frameworks.

\subsection{LightAutoML’s Tabular Preset}
TabularAutoML is the default LightAutoML pipeline that solves three types of tasks on tabular data: \emph{binary classification}, \emph{multiclass classification}, and \emph{regression}
for various types of loss functions and performance metrics. The input data for TabularAutoML is a table having columns of the following four types: numeric features, categorical features, timestamps, and a single target column with continuous values or class labels. 
\par
The key features of our LightAutoML pipeline are:
\begin{itemize}[leftmargin=*]
	\item \emph{Strong baseline}: works good on most datasets
	\item \emph{Fast}: no metamodels or pipeline optimization
	\item \emph{Advanced data preprocessing} compared to the other popular open source solutions
\end{itemize}
\par
One of our main goals in designing LightAutoML was to make a tool for fast hypothesis testing.
Therefore, we avoid using brute-force methods for optimal pipeline search and focus only on the models and efficient techniques that work across a wide range of datasets. In particular, we train \emph{only} two classes of models represented by three types of algorithms in the following order: \emph{Linear Model} with L2 penalty, the \emph{LightGBM} version of the GBM method~\cite{ke2017lightgbm} and the \emph{CatBoost} version of GBM~\cite{dorogush2018catboost}. 
\par
The selected order matters here because it helps to manage time if the user sets the time limit. The algorithm’s training order is ranked by the time they usually spend training. Therefore, if the time limit is small but reasonable, at least the fastest model will be computed. On the other hand, if the time limit is large enough, fast evaluation of the previous models allows us to allocate more time for training and hyperparameter tuning for the slower ones.
\par
Traditional ML algorithms were selected for the LightAutoML system because, despite the trend in the development of neural networks for different domains, GBM-based methods show strong performance results on tabular data and outperform other approaches in many benchmarks and competitions at the moment. Furthermore, various GBM frameworks are commonly used in industry to develop production models \cite{ke2017lightgbm, dorogush2018catboost, chen2016xgboost}. In addition, linear models are fast, easy to tune, and they can boost the performance of tree-based models in ensembles by adding a variety of predictions \cite{breiman1996bagging}. 
In comparison, other popular open source AutoML frameworks usually use significantly more classes of models and therefore require more time to train. As we show in Section~\ref{subsec:OpenMLBenchmark}, the proposed approach with a few additional features is able to outperform existing solutions on internal datasets used in our company and on the OpenML benchmark \cite{gijsbers2019open}.

\subsection{Data Preprocessing and Auto-typing}
\label{subsec:AutoType}

As we initially started developing LightAutoML, we paid special attention to the data preprocessing part. As already mentioned in Section~\ref{sec:intro}, we need to be ready to work with datasets in different formats, scales, containing artifacts, NaNs, or unspecified user processing. 
\par
To handle different types of features in different ways, we need to know each feature type. In the case of a single task with a small dataset, the user can specify each feature type manually. However, it becomes more problematic in the case of hundreds of tasks with datasets containing thousands of features. This is very typical for bank applications, and it takes hours of work of data scientists to perform this data analysis and labeling. So it is crucial for the AutoML framework to solve the problem of automatic data type inference \emph{(auto-typing)}.
\par
In case of TabularAutoML Preset we need to map features into three classes: \emph{numeric}, \emph{category}, and \emph{datetime}. One simple and obvious solution is to use column array data types as actual feature types, that is, to map float/int columns to numeric features, timestamp or string, that could be parsed as a timestamp --- to datetime, and others to category.
However, this mapping is not the best because of the frequent occurrence of numeric data types in category columns. An ML-based decision for this problem is described in \cite{shah2019ml}, where different models over meta statistics are used to predict human feature type labeling. Deep learning is also used to solve a similar but slightly different problem of semantic data type detection in \cite{hulsebos2019sherlock}.
\par
We solve this problem in a slightly different way. Let us say, 
column has category type if category encoding techniques, such as target encoder \emph{(OOFEnc)} \cite{micci2001preprocessing, dorogush2018catboost} or frequency encoder \emph{(FreqEnc)}, that encodes category by the number of occurrence in train sample, perform better than numeric ones, such as raw or discretized by quantiles \emph{(QDiscr)} values. Building a lot of models to verify the performance of all encoding combinations becomes impractical, so we need some proxy metric that is simple to calculate and agnostic for the type of LightAutoML task, including given loss and metric functions. We choose Normalized Gini Index \cite{chen1982gini} between target variable and encoded feature as a measure of encoding quality because it estimates the quality of sorting and could be computed for both classification and regression tasks. The details of the auto-typing algorithm are presented in Appendix~\ref{appendix:autotyp} (as Algorithm~\ref{algo:AutoTyping}). The final decision is made by ten expert rules over estimated encoding qualities and some other meta statistics such as the number of unique values. Exact list of rules is given at the LightAutoML repository\footnote{\href{https://github.com/sberbank-ai-lab/LightAutoML/blob/master/lightautoml/reader/guess\_roles.py}{https://github.com/sberbank-ai-lab/LightAutoML/blob/master/lightautoml/reader\\/guess\_roles.py}}. 
\par
Note that we do not use ML models for auto-typing because, as it was mentioned before, our goal is not to predict human labeling but to guess what is actually better for the final model performance.
Sometimes LightAutoML \emph{auto-typing} prediction may differ from human’s point of view, however it may lead to a significant boost in the performance, see Section~\ref{subsec:Ablation} and datasets guilermo, amazon\_employee, and robert in Table~\ref{OpenMLResults}
that contain a lot of categories 
from \emph{auto-typing} point of view.
\par
After we infer the type of a feature, we can additionally guess the optimal way to preprocess it, for example, to select the best category encoding or decide if we should discretize numbers. A similar algorithm can be used for this purpose with a small adaptation by using different rules and encoding methods. 

\subsection{Validation Schemes}

As was mentioned earlier, data in the industry may rapidly change over time in some ecosystem processes, which made independent identically distributed \emph{(IID)} assumption irrelevant in model development. There are cases when time series-based, grouped, or even custom validation splits are required. This becomes important because validation in AutoML is used not only for performance estimation but also for hyperparameters search and out-of-fold prediction generation. Out-of-fold prediction is used for blending and stacking models on upper LightAutoML levels and also returned as the prediction on train set for user analysis.
\par
To the best of our knowledge, other popular AutoML frameworks use only classical KFold or random holdout approaches, while advanced validation schemes help to handle non \emph{IID} cases and make models more robust and stable in time. This problem is out of the scope of OpenML benchmark tasks but becomes important in production applications. Currently available validation schemes in TabularAutoML are:
\begin{itemize}[leftmargin=*]
    \item \textbf{KFold cross-validation}, which is used by default (including stratified KFold for classification tasks or GroupKFold for behavioral models if group parameter for folds splitting is specified)
    \item \textbf{Holdout validation} if holdout set specified
    \item \textbf{Custom validation schemes} (including time series split \cite{cerqueira2020evaluating} cross-validation)
\end{itemize}
\par
All models that are trained during the cross-validation loop on different folds are then saved for the inference phase. Inference on new data is made by averaging models from all train folds. 

\subsection{Feature Selection}

Feature selection is a crucial part of industrial model development because it allows one to reduce model implementation and inference costs. However, existing open source AutoML solutions are not focused much on this problem. In turn, TabularAutoML implements three strategies of feature selection: \emph{No selection}, \emph{Importance cut off selection} (by default), \emph{Importance based forward selection}.
\par
Feature importance could be estimated in two ways: split-based tree importance \cite{lundberg2017unified} or permutation importance \cite{altmann2010permutation} of GBM model. Importance cutoff selection aims to reject only features that are useless for the 
model (importance measure <= 0). This strategy helps to reduce the number of features with no performance decrease, which may speed up model training and inference. 
\par
However, in order to reduce inference costs, one may want to limit the number of features in the model or to find a minimal possible model with a small performance drop. For that purpose, we implement a variant of the classical forward selection algorithm described in \cite{guyon2003introduction} with the key difference of ranking the candidate features by the importance measure mentioned above that helps to speed up the procedure significantly. The specifics of the algorithm are provided in Appendix~\ref{appendix:pbfs} (as Algorithm~\ref{FBFS}).
\par
We show in Table~\ref{Selection} on internal datasets that it is possible to build much faster and simpler models with slightly lower scores.

\begin{table}[htbp]
    \caption{Comparison of different selection strategies on binary bank datasets.}
    \label{Selection}
    \begin{tabular}{ccc}
    \toprule
    Strategy & Avg ROC-AUC & Avg Inference Time (10k rows) \\
    \midrule
Cut off & 0.804             & 9.1078              \\
Forward & 0.7978            & 5.6088                \\
    \bottomrule
\end{tabular}
\end{table}

\subsection{Hyperparameter Tuning}

In TabularAutoML, we use different ways to tune hyperparameters depending on what is tuned:
\begin{itemize}[leftmargin=*]
	\item \textbf{Early stopping} to choose the number of iterations (trees in GBM or gradient descent steps) for all models during the training phase
	\item \textbf{Expert system}
	\item \textbf{Tree structured Parzen estimation} \emph{(TPE)} for GBM models
	\item \textbf{Grid search}
\end{itemize}
\par
All hyperparameters are tuned by maximizing the metric function, which is the default for the solved task or which is defined by the user.

\subsubsection{Expert system}

A simple way to quickly set hyperparameters for models in a satisfactory \cite{simon1956rational} fashion is expert rules. TabularAutoML can initialize a \textquotedblleft reasonably good\textquotedblright\ set of GBM hyperparameters (learning rate, subsample, columns sample, depth) depending on a task and a dataset size. Suboptimal parameter choice is partially compensated by an adaptive selection of the number of steps using early stopping. It prevents the final model from a high decrease in score compared to hard-tuned models.

\subsubsection{TPE and combined strategy}
\label{subsubsec:Combined}

We introduce a mixed tuning strategy that works by default in TabularAutoML (but the user might change it): for each GBM framework (LightGBM and CatBoost) we train two types of models. The first one gets expert hyperparameters, and the second one is fine-tuned while it fits into the time budget. TPE algorithm, described in \cite{bergstra2011implementations} is used for the model fine-tuning. This algorithm is chosen because it shows state-of-the-art results in tuning this class of models. We use realization of TPE by Optuna framework \cite{akiba2019optuna}.
In the final model, both models will be blended or stacked. Also, one of the models (or even both) could be dropped from the AutoML pipeline if it does not help to increase the final model performance. We compare this combined strategy with the AutoML based on the expert system only; results are given in Section~\ref{subsec:Ablation}.

\subsubsection{Grid search}

Grid search parameter tuning is used in TabularAutoML pipeline to fine-tune regularization parameter of a linear model in combination with:
\begin{itemize}[leftmargin=*]
    \item \textbf{Early stopping}. We assume the regularization parameter in the linear model has a single optimal point, and after reaching it, we can finish the search.
    \item \textbf{Warm start}. We use warm start 
    strategy to initialize model weights between regularization trials \cite{friedman2010regularization}. It helps to speed up model training.
\end{itemize}
\par
Both heuristics make the grid search efficient in fine-tuninig linear estimators.

\subsection {Model Ensembling in TabularAutoML}

\subsubsection{Multilevel stacking ensembles}
\label{subsubsec:Stacking}

As was mentioned before, LightAutoML allows users to build stacked ensembles of unlimited depth. Similar strategy is common for AutoML systems and also used in \cite{ledell2020h2o, erickson2020autogluon}. However, building ensembles deeper than 3 levels shows no effect in practice. 
\par
TabularAutoML builds two-level stacking ensembles by default for multiclass classification tasks only because it was the only case where we observed a significant and stable boost in model performance; see Section~\ref{subsec:Ablation}. This behavior is just the default setting and can be changed by the user to perform stacking on any dataset.

\subsubsection{Blending}

The last level of LightAutoML ensemble, regardless of its depth, may contain more than one model. To combine predictions of different models into a single AutoML output, they are passed to the blending phase. Blending in terms of LightAutoML differs from the full stacked model in the following ways. First, it is a considerably simpler model. As a consequence, second, it does not require any validation scheme to tune and control overfitting. Finally, it is able to perform model selection to simplify the ensemble to speed up the inference phase.
\par
TabularAutoML uses weighted averaging as a blender model. Ensemble weights are estimated using the coordinate descent algorithm to maximize the metric function, which is the default for the solved task or which is defined by the user. Models with weights close to $0$ are dropped from AutoML.

\subsubsection{Ensemble of AutoMLs and time utilization}
\label{subsubsec:Utilization}

As we mentioned before, one of our goals was to limit the search space of ML algorithms to speed up model training and inference, which is related to production use cases of AutoML frameworks for medium and high dataset sizes. However, this strategy will not perform well in cases such as ML competitions when the user has a very high time budget and needs to utilize it all to get the best performance regardless of its cost. Increasing the search space and brute forcing will often take advantage here.
\par
Typical examples here are small datasets from the OpenML benchmark that are solved by TabularAutoML much faster than the given time limit of 1 hour (Table~\ref{UtilTable}). 
In order to solve this problem and be competitive on benchmarks and ML competitions in the case of small datasets, we implement the time utilization strategy that blends multiple TabularAutoMLs with slightly different settings and validation random seeds. A user may define multiple config settings and the priority order or use defaults. AutoMLs with the same settings and different seeds will be simply averaged together, and after that, different settings ensembles will be weighted averaged. This strategy shows a performance boost on OpenML tasks; see Section~\ref{subsec:Ablation}.

\begin{table}[htbp]
    \caption{Average training time in seconds for the smallest OpenML datasets for Utilized Tabular Preset version and Default (single run).}
    \label{UtilTable}
    \begin{tabular}{ccc}
    \toprule
    Task type & Utilized & Single run \\
    \midrule
Binary (9 smallest)      & 3268     & 360        \\
Multi class (7 smallest) & 2984     & 1201        \\
    \bottomrule
\end{tabular}
\end{table}

\section{Performance of LightAutoML}

\subsection{Comparison with Open Source AutoML}
\label{subsec:OpenMLBenchmark}

In this section, we compare the performance of LightAutoML TabularAutoML Preset with the already existing open source solutions across various tasks and show the superior performance of our method. First, to make this comparison, we use datasets from the OpenML benchmark that is typically used to evaluate the quality of AutoML systems. The benchmark is evaluated on $35$ datasets of binary and multiclass classification tasks. The complete experimental description, including the framework versions, limitations, and extended results, is presented in Appendix~\ref{OpenMLDetails}. The summary of the results of LightAutoML vis-a-vis five popular AutoML systems is presented in Table~\ref{AggComp}, where all the AutoML systems are ranked by the total amount of wins in each dataset. 

\begin{table}[htbp]
    \caption{Aggregated framework comparison on OpenML.}
    \label{AggComp}
    \begin{tabular}{cccc}
    \toprule
    framework   & Wins & Avg Rank & Avg Reciprocal Rank \\
    \midrule
autoweka    & 0    & 5.7879   & 0.1747              \\
autogluon   & 0    & 4.2647   & 0.252               \\
autosklearn & 3    & 2.6      & 0.4505              \\
tpot        & 6    & 3.8235   & 0.374               \\
h2oautoml   & 6    & 2.4857   & 0.4833              \\
lightautoml & \textbf{20}   & \textbf{1.9429}   & \textbf{0.7233}             \\
    \bottomrule
\end{tabular}
\end{table}

\par
However, the detailed comparison of frameworks in the context of dataset groups provided in Table~\ref{OpenMLDetailedRecip} shows that LightAutoML does not work equally well on all the classes of tasks. For binary classification problems with a small amount of data, LightAutoML shows average performance results and losses to TPOT; moreover, it performs on par with H2O and Auto-Sklearn. The reason for this is that the tasks with a small amount of data are not common in our ecosystem and were not the main impetus behind the development of LightAutoML.

\begin{table}[htbp]
    \caption{Average reciprocal rank for frameworks comparison by OpenML datasets groups.}
    \label{OpenMLDetailedRecip}
    \begin{tabular}{ccccc}
    \toprule
     & Small  & Small  & Medium  & Medium  \\    
Framework       &  binary &  multiclass &  binary &  multiclass \\
    \midrule
autoweka    & 0.1741       & 0.1667           & 0.1783        & 0.1786            \\
tpot        & \textbf{0.6056}       & 0.481            & 0.2061        & 0.2333            \\
autogluon   & 0.2796       & 0.2071           & 0.2606        & 0.2476            \\
autosklearn & 0.4519       & 0.3571           & 0.4424        & 0.5417            \\
h2oautoml   & 0.4907       & 0.5595           & 0.4697        & 0.4271            \\
lightautoml & 0.4481       & \textbf{0.6786}           & \textbf{0.8939}        & \textbf{0.8375}           \\
    \bottomrule
\end{tabular}
\end{table}

\par
Another type of datasets for comparing different solutions is internal datasets collected in the bank. In this study, we use 15 bank datasets for various binary classification tasks performed in our company, such as credit scoring (probability of defaults estimation), collection, and marketing (response probability). As the primary goal of developing the LightAutoML framework was to work with our internal applications, we expected better performance of our system on the internal data. In Table~\ref{AggCompInner} we show that the performance gap between LightAutoML and other AutoML systems is significantly higher on the bank datasets than on the OpenML data\footnote{We cannot present information about these internal datasets in this paper because they contain proprietary and confidential information.}.

\begin{table}[htbp]
    \caption{Aggregated framework comparison on bank's proprietary datasets.}
    \label{AggCompInner}
    \begin{tabular}{cccc}
    \toprule
    framework   & Wins & Avg Rank & Avg Reciprocal Rank \\
    \midrule
autoweka    & 0    & 5.7333   & 0.18              \\
autogluon   & 0    & 3.9333   & 0.2778           \\
tpot        & 1    & 3.9333   & 0.3056           \\
autosklearn & 1    & 3.2667      &  0.3956          \\
h2oautoml   & 1    & 2.6667   & 0.4244              \\
lightautoml & \textbf{12}   & \textbf{1.4667}   & \textbf{0.8667}             \\
    \bottomrule
\end{tabular}
\end{table}

\subsection{Ablation Study}
\label{subsec:Ablation}

We perform the ablation study estimating average reciprocal rank change to evaluate each TabularAutoML feature impact on the OpenML benchmark results. We take the best existing AutoML configuration, including the time utilization strategy, combined hyperparameter tuning, auto-typing, stacking for multiclass tasks only \textit{(Utilized best)} as baselines. First, we turn off time utilization to estimate contribution of multi-start bagging \textit{(Default)} described in Section~\ref{subsubsec:Utilization}. Second, we take \textit{Default} and compare the two-level stacking for all the tasks \textit{(Stacked all)} and blending first level models only \textit{(Single level all)} to estimate the quality of alternative ensembling methods discussed in Section~\ref{subsubsec:Stacking}. Third, we exclude the advanced auto-typing module presented in Section~\ref{subsec:AutoType} from \textit{Default} \textit{(No auto-typing)}. Finally, we replace the combined tuning strategy described in Section~\ref{subsubsec:Combined} from \textit{Default} with the expert system initialization only \textit{(No finetune)}. The ablation study results are presented in Table~\ref{Abligation}. As Table~\ref{Abligation} demonstrates, each feature removal decreases the LightAutoML rank, which shows that all those features make our framework more accurate than the others on OpenML datasets. 

\begin{table}[htbp]
    \caption{Ablation study on OpenML.}
    \label{Abligation}
    \begin{tabular}{ccc}
    \toprule
    Configuration                & Avg Reciprocal Rank & Avg Rank \\
    \midrule
No finetune                  & 0.6054              & 2.4118   \\
Typing off                   & 0.6431              & 2.1765   \\
Single level all             & 0.65                & 2.2353   \\
Stacked all                  & 0.6672              & 2.2059   \\
Default (stacked multiclass) & 0.6907              & 2.0882   \\
Utilized best                & \textbf{0.7233}              & \textbf{1.9429}    \\
    \bottomrule
\end{tabular}
\end{table}

\subsection{LightAutoML vs. Building Models by Hand}
\label{subsec:NextHack}

We have also used our LightAutoML system as one of the 
\textquotedblleft participants\textquotedblright\ in the 
internal hackathon in our ecosystem, together with $433$ leading data scientists in the company. The training dataset used in the competition had $300$ features and $400\,000$ records, and the goal was to predict the churn rates. The performance metric selected for this hackathon was ROC-AUC, and the performance of the baseline used in this competition was ROC-AUC = $75.5\%$. 
\par
\begin{figure}[htbp]
  \centering
  \includegraphics[width=0.95\linewidth]{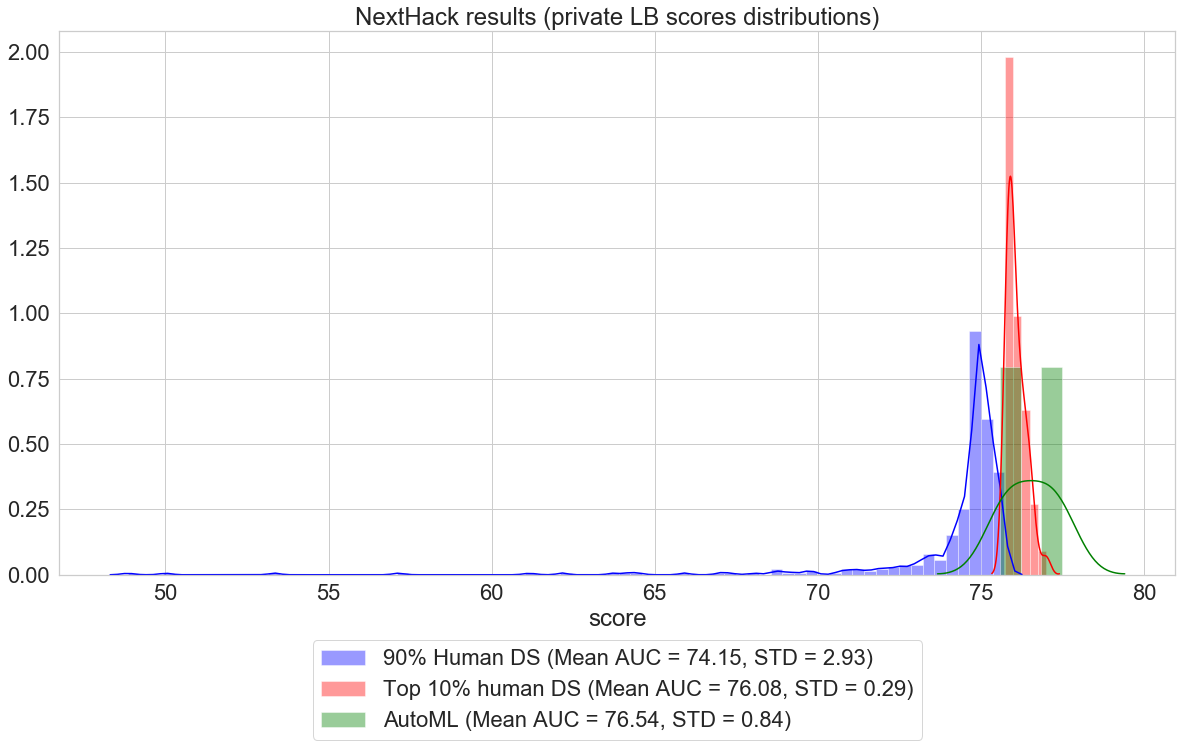}
  \caption{Performance Results of LightAutoML in the NextHack Competition vs. 433 Human Competitors.}
\label{nexhack_lb}
\end{figure}

LightAutoML was presented in the hackathon by $4$ participants that used it in different configurations. As Figure~\ref{nexhack_lb} demonstrates, LightAutoML outperformed the baseline model. Although the average performance of LightAutoML (ROC-AUC = 76.54) was better than the average performance of the top-10\% of hackathon participants, i.e., belonging to the $90\%$ quantile (average ROC-AUC = 76.08), the performance improvements were not statistically significant. This means that although LightAutoML significantly outperformed the average data scientist in this hackathon (including the \textquotedblleft bottom $90\%$\textquotedblright), its performance was comparable to the top-$10\%$ of the best data scientists. Detailed results are given in repository\footnote{\url{https://github.com/sberbank-ai-lab/Paper-Appendix}}.

\section{Deploying LightAutoML} 

In this section, we present our experiences with deploying, piloting, and moving LightAutoML into production.
\par
\textbf{Deployment}.
\label{subsec:deployment}
Currently, the LightAutoML system is deployed in production by five large ML platforms inside our financial services company and its ecosystem, including cloud, B2B, and B2C platforms. Furthermore, seven more divisions are currently piloting the latest version of the system. Moreover, it is also used in several automated systems and various IT services across the ecosystem. Altogether, more than $70$ teams involving several hundred data scientists use LightAutoML to build ML models across the entire ecosystem. As an example, just the B2C platform alone has more than $300$ business problems solved using LightAutoML this year, resulting in the total P\&L increase by $3\%$. In what follows, we present some examples of successful deployments of LightAutoML in the ecosystem and our experiences with these deployments.
\par
\emph{Operational audit.}
LightAutoML has been applied to the problem of operational audit of the bank branches with the goal to detect mistakes made by bank's employees across the organization and do it in the most effective and efficient manner. These mistakes are of numerous types, depending on the type of branch, its location, the type of employee who made a mistake, etc. The goal of the operational audit is to detect and correct all these mistakes, prevent their future occurrences, and minimize their consequences according to the established practices of the bank.

\begin{figure*}[htbp]
  \centering
  \includegraphics[width=\linewidth]{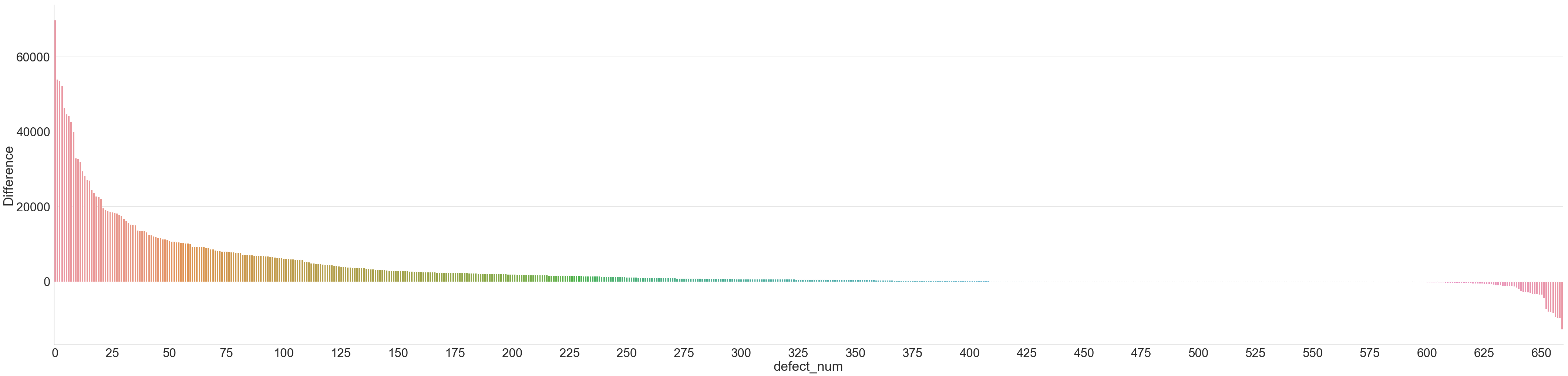}
  \caption{Difference in revenues of the LightAutoML and the manually developed impact estimation methods.}
  \label{Fig3}
\end{figure*}
\par
In this project, we focused on 60 major types of mistakes and developed one predictive model per mistake and each of the $11$ divisions of the bank, resulting in $660$ LightAutoML models in total. For comparison, prior mistake detection methods were rule-based. 
\par
One of the reasons for the bank not developing prior ML-based mistake detection models was the large number of such models ($660$ in our case) requiring extensive resources of data scientists and a long time to produce all of them (measured in person-years), making such project infeasible. The second reason why such models have not been previously developed is that the economic effect from each model is limited. For example, Figure~\ref{Fig3} shows the $660$ operational risk models sorted on the $y$-axis by the difference in the economic impact between the LightAutoML models and the previously existing rule-based methods, ranging from $\$60\,000$ highest positive impact on the left to the $\$8\,000$ highest negative impact on the right of Figure~\ref{Fig3}. This example demonstrates that detecting even the most important operational mistakes resulted in limited savings, and the savings from the medium and minor mistakes were considerably lower, making the development of such machine learning models economically infeasible. 
Nevertheless, the \emph{cumulative} economic effect of building \emph{all} the $660$ models is significant, bringing the bank millions of dollars in savings. 
\par
All the $660$ operational audit models were developed by our LightAutoML system in $3$ days over the weekend, and the whole project took $10$ person-days, taking into account data preparation costs. This contrasts sharply with the cost of creating so many models manually, resulting in saving the bank millions of dollars. 
\par
\emph{Other Examples of LightAutoML Applications.} LightAutoML system has also automatically built fraud detection models that were subsequently compared against the manually developed classical ML-based fraud detection models previously developed and deployed in the company. These models saved development time by $40$ person-days and improving model performance by $6\%$ for the F1 performance metric, identifying several thousands of fraudulent activities, and thus saving the bank millions of dollars.
\par
Another example of successful LightAutoML deployment was a charitable donations system involving $110$ different types of contributions focusing on children's welfare. 
Our LightAutoML system developed a model identifying the donors over a period of two person-days. When it was moved into production, only for the email channel it increased the number of donations by $18\%$ and the total sum of donations by $40\%$.
\par
\textbf{Lessons Learned}. 
\label{Lessons learned}
We will start with the lessons learned when \emph{developing and piloting} LightAutoML for our ecosystem.
First, applications that are the most amenable to the deployment of AutoML are those where the prediction problem is formulated precisely and correctly and does not require human insights based on deep prior knowledge of the application domain. 
Second, AutoML is well suited for typical supervised learning tasks, where testing works well on retrospective data. It is harder to effectively apply the current generation of AutoML systems to non-standard problems. 
Third, AutoML solutions are especially useful when an ML model should be built quickly with limited resources. In those cases when there is plenty of time to build a model using top data science talent working on complex problems requiring significant insights and ingenuity, non-standard solutions, and careful fine-tuning of the model parameters, humans can outperform AutoML solutions. As an example, when Google conducted the IEEE competition \cite{Rishi2019}, its AutoML solution outperformed 90\% of the competing teams in the first two weeks. However, humans eventually managed to catch up and outperform Google’s AutoML after the first two weeks by putting extensive effort into the hackathon and constantly improving their models.
Finally, strong performance results of LightAutoML were achieved not due to certain unique breakthrough features of our system but because of numerous incremental improvements described in Section~\ref{sec:lama} that were cohesively combined into a unified solution.
\par
Furthermore, while moving LightAutoML into \emph{production},
we have learned a different group of lessons.
First, independent tests of our system by data scientists in the company show that LightAutoML significantly outperformed humans on only one-third of the ML tasks actually deployed in production, which differs from the $90\%$ figure reported in Section~\ref{subsec:NextHack} for the pilot studies. This reduction of LightAutoML performance vs. humans in production environments is due to the following reasons:
\begin{itemize}[leftmargin=*]
     \item Presence of \emph{data leaks}, i.e., the situation when information about an object target becomes obvious from features for the training and not for the test data, and also of \emph{data artifacts}, such as special decimal separators or character symbols (e.g., K standing for thousands)
     in numeric columns, missing values for a specific feature block, 
     etc., across hundreds of automated data storage systems.
    \item Existence of extremely small number of minority class events, compared with the number of features.
     \item In the same conditions, top data scientists build many more models than less experienced ones at the same time.
\end{itemize}
\par
The last point demonstrates that, instead of replacing data scientists with AutoML systems, it is better to complement them with such systems. Now, in our company, we focus on empowering our DSes with LightAutoML. 
In particular, we are primarily using LightAutoML 
as a baseline generator and as a fast hypothesis testing tool in the company now. This helps our data scientists to focus on certain crucial parts of the model development process, such as appropriate time period data selection, target variable formulation, selection of suitable quality metrics, identification of business constraints, and so on. 
\par
The second lesson is associated with the importance of integrating LightAutoML with different production environments of our diverse ecosystem to implement end-to-end solutions.
Although we observed 4x to 10x model training time reduction in comparison to the usual model creation process, overall time-to-market managed to decrease by only $30\%$ on average for the \emph{whole} model life-cycle process. 
Furthermore, we observed that this number can be improved 
to almost $70\%$ for the cases when continuous integration with data sources and inference environments was done, notably on our cloud platform.
\par
In summary, we had encountered several issues when piloting our LightAutoML system and moving it into production, most of these issues have to deal with idiosyncratic requirements of the financial services industry and the diverse ecosystem of our organization. We managed to resolve them successfully by developing the LightAutoML system to suit the needs of our ecosystem.
We have also described the lessons learned while moving LightAutoML into production across a diverse class of ML applications.
All this makes a strong case for developing vertical AutoML solutions for the financial services industry.

\section{Conclusion}

In the paper, we present the LightAutoML system designed to satisfy the specific needs of large financial services companies and their ecosystems. We argue for the need to develop a special-purpose AutoML, as opposed to the general-purpose system, such as H2O or AutoGluon, that would satisfy the idiosyncratic needs of such organizations, including the ability to handle large datasets having a broad range of data types, non-stationary data, specific types of validations, including behavioral models and out-of-time validations, and rapid development of a large number of models. 
The proposed LightAutoML system has several incremental improvements: the \textquotedblleft light-and-fast\textquotedblright\ approach to AutoML development when only GBMs and linear models are used, novel and fast combined hyperparameter tuning method that produces strong tuning results, advanced data preprocessing including auto-typing that collectively enhances functionality of LightAutoML and helps it to achieve superior performance results.
\par
Further, we show that our LightAutoML system outperforms some of the leading general-purpose AutoML solutions in terms of the AUC-ROC and LogLoss metrics on our proprietary applications and also on the OpenML benchmarks, as well as the models manually developed by data scientists for the typical problems of importance to  large financial organizations.  
\par
Finally, the proposed LightAutoML system has been deployed in production  in numerous applications across the company and its ecosystem, which helped to save the organization millions of dollars in development costs, while also achieving certain capabilities that are impossible for the humans to realize, such as the generation of massive amounts of ML models in record time. We have also described several important lessons that we have learned while developing and deploying the LightAutoML system at the company, including that (a) the \textquotedblleft light\textquotedblright\ approach to AutoML design in the LightAutoML system worked well in practice, achieving superior performance results --- mainly due to the careful integration of various incremental improvements of different AutoML features properly combined into the unified LightAutoML system; (b) realization that LightAutoML outperformed data scientists on only one third of the deployed models, as opposed to the expected $90\%$ of the cases --- due to the complexities and the \textquotedblleft messiness\textquotedblright\ of the actually deployed  vis-a-vis the pilot cases; (c) realization that it is not always true that data scientists outperform the machines when preparing the data to be used for building ML models --- LightAutoML outperformed data scientists in this data preparation task in several use cases; (d) although LightAutoML significantly improved model building productivity in our organization, the number of data scientists actually increased significantly over the last year~---~mainly due to the fact that we need many more ML models to better run our business. There is plenty of work for both AutoML and data scientists to achieve our business goals.
\par
As a part of the future work, we plan to develop functionality related to model distillation and strengthen the work with NLP tasks.
In particular, some applications in our organization, including e-commerce, impose additional constraints on the real-time performance of ML models, and we need the make sure that the distillation component of LightAutoML satisfies these real-time requirements. 
Furthermore, we plan further to enhance NLP functionality of LightAutoML in its future releases.

\bibliographystyle{ACM-Reference-Format}
\bibliography{ref}

\clearpage
\appendix

\section{Experiment design and additional results}
\label{OpenMLDetails}

Datasets and experiment design were taken from the official OpenML benchmark page. All the datasets were evaluated across 10 cross-validation folds made by organizers, and the final score for each dataset was calculated by averaging scores from all 10 folds. Models were scored by ROC-AUC metric for binary classification tasks and LogLoss for multiclass classification tasks. In this paper, we drop from evaluation 4 out of 39 datasets because most of the frameworks failed on these datasets due to timeout with given limitations.
\par
Models were evaluated with the following limitations: 1 hour runtime limit (this limit was passed to the framework as the input parameter if the framework supports time limitations, but actually process was killed after 2 hours), 8 CPU, 32 GB RAM per single cross-validation split. Each fold was evaluated in a separated docker container on a cloud server under OS Ubuntu 18.04 with HDD and Intel(R) Xeon(R) Gold 6148 CPU @ 2.40GHz. 
\par
Frameworks and versions that were compared in the benchmark are the following: lightautoml==0.2.8, h2o\_automl==3.32.0.1, autogluon==0.0.12, autosklearn==0.11.1, autoweka==2.6, tpot==0.11.5. The code for benchmark evaluation was taken from the OpenML repository\footnote{\url{https://github.com/openml/automlbenchmark}}, were it was published by frameworks developers. Finally, we have only 4 failure cases due to timeout. Results across all 35 datasets are shown in Table~\ref{OpenMLResults}.
\par
The code to reproduce our experiment is available at the repository\footnote{\url{https://github.com/sberbank-ai-lab/automlbenchmark/tree/lightautoml}}. The code to reproduce LightAutoML results is also published at the OpenML repository.
\par
Note that, as it is mentioned in \cite{erickson2020autogluon}, AutoGluon shows state-of-the-art results on the OpenML benchmark, but we were not able to reproduce the results using the code published at the OpenML repository. A possible reason for this is non-default run settings or differences in computational environments.
Also, during the benchmarks process, we encountered various errors in the tested frameworks; most of them are listed in Appendix in \cite{erickson2020autogluon}. 
\par
The comparison for the internal datasets was made on the same environment except the time limit --- we set the limit to two hours for inner data. Datasets contain clients' information that can not be published, so only aggregated framework comparison is presented.
Datasets were split independently into train/test samples by data owners, depending on their business tasks. The split may be random, out-of-time, or separated by group values (for example, by client IDs for behavioral models). Split methods and test samples were unknown to AutoMLs during the training phase. Models were trained on train parts and scored by ROC-AUC metric values on test parts.

\section{Auto-typing algorithm}
\label{appendix:autotyp}
\removelatexerror
\begin{algorithm}[H]
    \SetAlgoLined
    \LinesNumbered
    \SetKwInOut{Input}{Input}
    \SetKwInOut{Output}{Output}

    \Input{Integer and float type train features $X_{train}^{if}$, train target $Y_{train}$, set of expert rules $Rules$}
    
    \Output{Boolean values for each feature in $X_{train}^{if}$ to be numeric $IsNumber$}
    
    \caption{Splitting integer and float features into categorical and numeric features.}
    \label{algo:AutoTyping}
    
    \For{each feature $F$ in $X_{train}^{if}$}{
        $IsNumber[F] = False$
    
        $notNaNSlice = isNotNaN(F)$
        
        $F^{nn} = F[notNaNSlice]$
        
        $Y_{train}^{nn} = Y_{train}[notNaNSlice]$
        
        $NG_{noEnc} = NormGini(Y_{train}^{nn}, F^{nn})$\; 
        
        $NG_{OOFq} = NormGini(Y_{train}^{nn}, OOFEnc(QDiscr(F^{nn})))$\;
        
        $NG_{FE} = NormGini(Y_{train}^{nn}, FreqEnc(F^{nn}))$\;
        
        $NG_{OOF} = NormGini(Y_{train}^{nn}, OOFEnc(F^{nn}))$\;
        
        \For{$R$ in $Rules$}{
            \If{$R(F^{nn}, NG_{noEnc}, NG_{OOFq}, NG_{FE}, NG_{OOF})$}{
                $IsNumber[F] = True$
            
                Break\;
            }
        }
        
    }
\end{algorithm}

\section{Permutation based forward selection algorithm}
\label{appendix:pbfs}
\removelatexerror
\begin{algorithm}[H]
    \SetAlgoLined
    \LinesNumbered
    \SetKwInOut{Input}{Input}
    \SetKwInOut{Output}{Output}

    \Input{Train features $X_{train}$, train target $Y_{train}$, valid features $X_{valid}$, valid target $Y_{valid}$, MLAlgo $A$, features block size $N$, metric $M$}
    
    \Output{Selected features $OutFeats$}
    
    \caption{Importance based forward selection.}
    \label{FBFS}
    
    $T_{perm} = A(X_{train}, Y_{train})$\;
    
    $PFimp = PermutationImp(X_{valid}, T_{perm}, Y_{valid}, M)$\;
    
    Sort $FeatureNames$ by descending $PFimp$\;
    
    $OutFeats = []$\; 
    
    $BaselineScore = -inf$\;
    
    \For{$i$ from $0$ to $len(Features)$ with step $N$}{
        $CandidateFeats = FeatureNames[i: i + N]$\; 
        
        $T_{i} = A(X_{train}[OutFeats + CandidateFeats])$\; 
        
        $NewScore = M(Y_{valid}, T_{i}(X_{valid}))$\; 
        
        \If{$NewScore > BaselineScore$}{
             Append $CandidateFeats$ to $OutFeats$\;
            $BaselineScore = NewScore$\;
        }
    }
\end{algorithm}

\begin{table*}[htbp]
    \caption{Detailed results for OpenML datasets.}
    \label{OpenMLResults}

    \begin{tabular}{cccccccc}
    \toprule
    dataset            & metric  & lightautoml & autogluon & h2oautoml & autosklearn & autoweka & tpot    \\
    \midrule
australian         & roc-auc & \textbf{0.9462}      & 0.9393    & 0.934     & 0.9353      & 0.9337   & 0.9336  \\
blood-transfusi…   & roc-auc & 0.7497      & 0.719     & \textbf{0.758}     & 0.75        & 0.7282   & 0.7479  \\
credit-g           & roc-auc & 0.7921      & 0.7766    & \textbf{0.7968}    & 0.7756      & 0.7526   & 0.7824  \\
kc1                & roc-auc & 0.8283      & 0.8168    & 0.8374    & 0.8404      & 0.8166   & \textbf{0.844}   \\
jasmine            & roc-auc & 0.8806      & 0.8822    & 0.887     & 0.8826      & 0.8638   & \textbf{0.8897}  \\
kr-vs-kp           & roc-auc & 0.9997      & 0.9994    & 0.9997    & \textbf{0.9999}      & 0.981    & 0.9998  \\
sylvine            & roc-auc & 0.9882      & 0.9852    & 0.9882    & 0.9896      & 0.9729   & \textbf{0.9923}  \\
phoneme            & roc-auc & 0.9655      & 0.9682    & 0.9668    & 0.9634      & 0.9552   & \textbf{0.9693}  \\
christine          & roc-auc & \textbf{0.8307}      & 0.8133    & 0.8247    & 0.8285      & 0.7905   & 0.8065  \\
guillermo          & roc-auc & \textbf{0.9322}      & 0.9027    & 0.9078    & 0.9064      & 0.8901   & 0.8943  \\
riccardo           & roc-auc & 0.9997      & 0.9997    & 0.9997    & \textbf{0.9998}      & 0.9981   & 0.9906  \\
amazon\_employee…  & roc-auc & \textbf{0.9003}      & 0.8758    & 0.8756    & 0.8524      & 0.8363   & 0.8674  \\
nomao              & roc-auc & \textbf{0.9976}      & 0.9954    & 0.9959    & 0.9958      & 0.9826   & 0.9948  \\
bank-marketing     & roc-auc & \textbf{0.9401}      & 0.9371    & 0.9373    & 0.938       & 0.8103   & 0.9314  \\
adult              & roc-auc & \textbf{0.9306}      & 0.9286    & 0.9295    & 0.93        & 0.914    & 0.925   \\
kddcup09\_appete…  & roc-auc & \textbf{0.8509}      & 0.7932    & 0.8305    & 0.8383      & -        & 0.8111  \\
apsfailure         & roc-auc & \textbf{0.9936}      & 0.9915    & 0.9924    & 0.9921      & 0.9678   & 0.9904  \\
numerai28.6        & roc-auc & 0.5306      & 0.5212    & \textbf{0.5311}    & 0.5294      & 0.5249   & 0.5235  \\
higgs              & roc-auc & \textbf{0.8157}      & 0.8055    & 0.815     & 0.8137      & 0.676    & 0.8024  \\
miniboone          & roc-auc & \textbf{0.9876}      & 0.9842    & 0.9862    & 0.9865      & 0.9651   & 0.982   \\
car                & -logloss & -0.0038     & -0.1337   & -0.0033   & -0.0023     & -0.2216  & \textbf{-0.0001} \\
cnae-9             & -logloss & -0.1555     & -0.2917   & -0.2002   & -0.1784     & -0.8589  & \textbf{-0.1448} \\
connect-4          & -logloss & \textbf{-0.3358}     & -0.4956   & -0.3387   & -0.3481     & -3.1088  & -0.4411 \\
dilbert            & -logloss & \textbf{-0.0327}     & -0.1473   & -0.0411   & -0.0431     & -0.2455  & -0.0772 \\
fabert             & -logloss & -0.765      & -0.7715   & \textbf{-0.764}    & -0.7705     & -7.4649  & -0.8263 \\
fashion-mnist      & -logloss & \textbf{-0.2519}     & -0.3321   & -0.3025   & -0.2852     & -0.6423  & -0.3726 \\
helena             & -logloss & \textbf{-2.5548}     & -         & -2.8187   & -2.6359     & -15.3287 & -       \\
jannis             & -logloss & \textbf{-0.6653}     & -0.7275   & -0.7232   & -0.6696     & -4.1177  & -0.7319 \\
jungle\_chess\_2p… & -logloss & \textbf{-0.1428}     & -0.381    & -0.2366   & -0.1599     & -2.0565  & -0.2829 \\
mfeat-factors      & -logloss & \textbf{-0.0823}     & -0.1563   & -0.0941   & -0.093      & -0.542   & -0.1022 \\
robert             & -logloss & \textbf{-1.3166}     & -1.6828   & -1.6829   & -1.6571     & -        & -1.9923 \\
segment            & -logloss & \textbf{-0.0464}     & -0.0854   & -0.0497   & -0.0615     & -0.4275  & -0.0522 \\
shuttle            & -logloss & -0.0008     & -0.0008   & \textbf{-0.0005}   & -0.0005     & -0.0059  & -0.0006 \\
vehicle            & -logloss & -0.3723     & -0.4812   & \textbf{-0.3584}   & -0.3816     & -2.5381  & -0.3745 \\
volkert            & -logloss & -0.8283     & -0.9197   & -0.8669   & \textbf{-0.8054}     & -1.7296  & -0.9945\\

    \bottomrule
\end{tabular}
\end{table*}

\end{document}